\pdfoutput=1

\documentclass[11pt]{article}

\usepackage[preprint]{acl}

\usepackage{times}
\usepackage{latexsym}
\usepackage{listings}   
\usepackage{tcolorbox}  
\tcbuselibrary{listings, skins}  
\usepackage[T1]{fontenc}

\usepackage[utf8]{inputenc}

\usepackage{microtype}

\usepackage{inconsolata}

\usepackage{microtype}
\usepackage{booktabs} 
\usepackage{hyperref}
\usepackage{amsmath}
\usepackage{amssymb}
\usepackage{mathtools}
\usepackage{amsthm}
\usepackage{amsmath}
\usepackage{amssymb}
\usepackage{mathtools}
\usepackage{amsthm}
\usepackage{makecell}
\usepackage{bbm}
\usepackage[textsize=tiny]{todonotes}
\usepackage[capitalize,noabbrev]{cleveref}
\usepackage{inconsolata}
\usepackage{pythonhighlight}
\usepackage{algorithm2e}
\usepackage{utfsym}
\usepackage{fontawesome}
\usepackage{multirow}
\usepackage{booktabs}
\usepackage{adjustbox}
\usepackage{enumitem}
\usepackage{graphicx}
\usepackage{subfigure}
\usepackage{xcolor}
\usepackage{colortbl}
\usepackage{xcolor}
\usepackage{pgf}
\usepackage{pgfplotstable} 
\usepackage{pgfplots} 
\usepackage{colortbl} 
\usepackage{booktabs} 
\usepackage{etoolbox}
\usepackage{makecell}
\usepackage{courier}
\usepackage{authblk}
\usepackage{twemojis}
\newcommand{\sysname}{$\textit{\textbf{R}}^3$}

\title{\sysname~: ``This is My SQL, Are You With Me?'' A Consensus-Based Multi-Agent System for Text-to-SQL Tasks}

\author{
{\bf Hanchen Xia$^{*{\twemoji{peach}\twemoji{lemon}}}$, Feng Jiang$^{*\twemoji{lemon}}$, Naihao Deng$^{\twemoji{avocado}}$, Cunxiang Wang$^{\twemoji{lemon}}$, Guojiang Zhao$^{\twemoji{blueberries}}$}\\
{\bf Rada Mihalcea$^{\twemoji{avocado}}$, Yue Zhang$^{\twemoji{lemon}}$}\\
$^{\twemoji{peach}}$ School of Mathematical  Science, Shanghai Jiao Tong University\\
$^{\twemoji{lemon}}$ School of Engineering, Westlake University\\
$^{\twemoji{avocado}}$ University of Michigan  \quad$^{\twemoji{blueberries}}$ Carnegie Mellon University}

\begin{document}
\maketitle

\begin{abstract}
Large Language Models (LLMs) have demonstrated strong performance on various tasks.
To unleash their power on the Text-to-SQL task, we propose \sysname~(Review-Rebuttal-Revision), a consensus-based multi-agent system for Text-to-SQL tasks. 
\sysname~outperforms the existing single LLM Text-to-SQL systems as well as the multi-agent Text-to-SQL systems by $1.3\%$ to $8.1\%$ on Spider and Bird. 
Surprisingly, we find that for Llama-3-8B, \sysname~outperforms chain-of-thought prompting by over 20\%, even outperforming GPT-3.5 on the development set of Spider.
\end{abstract}

\section{Introduction}
Text-to-SQL, the task of converting natural language to SQL queries, enables non-technical users to access databases with natural language \cite{deng-etal-2022-recent, katsogiannis2023survey}. 
Recently, Large Language Models (LLMs) have made significant progress on various tasks \cite{touvron2023llama,openai2023gpt}.

Although researchers have proposed various methods to enhance the reasoning abilities of LLMs \cite{wei2022chain,yao2023tree,besta2024graph}, However, they are still facing challenges with Text-to-SQL tasks \cite{li2023can,hong2024next}. 
The LLM-based multi-agent system leverages collective intelligence from a group of LLMs and have achieved exceptional performance across various tasks \cite{park2023generative,hong2023metagpt,xu2023language}, but little work explores using them on Text-to-SQL.
The existing multi-agent Text-to-SQL system first decomposes the task into multiple subtasks which are then accomplished step-by-step by agents \cite{wang2023mac}. 
While achieving remarkable performances, such a decomposition-based system necessitates extensive manual prompt engineering and logic design. 

\begin{figure}[h!]
\centering
    \includegraphics[width=.75\columnwidth]{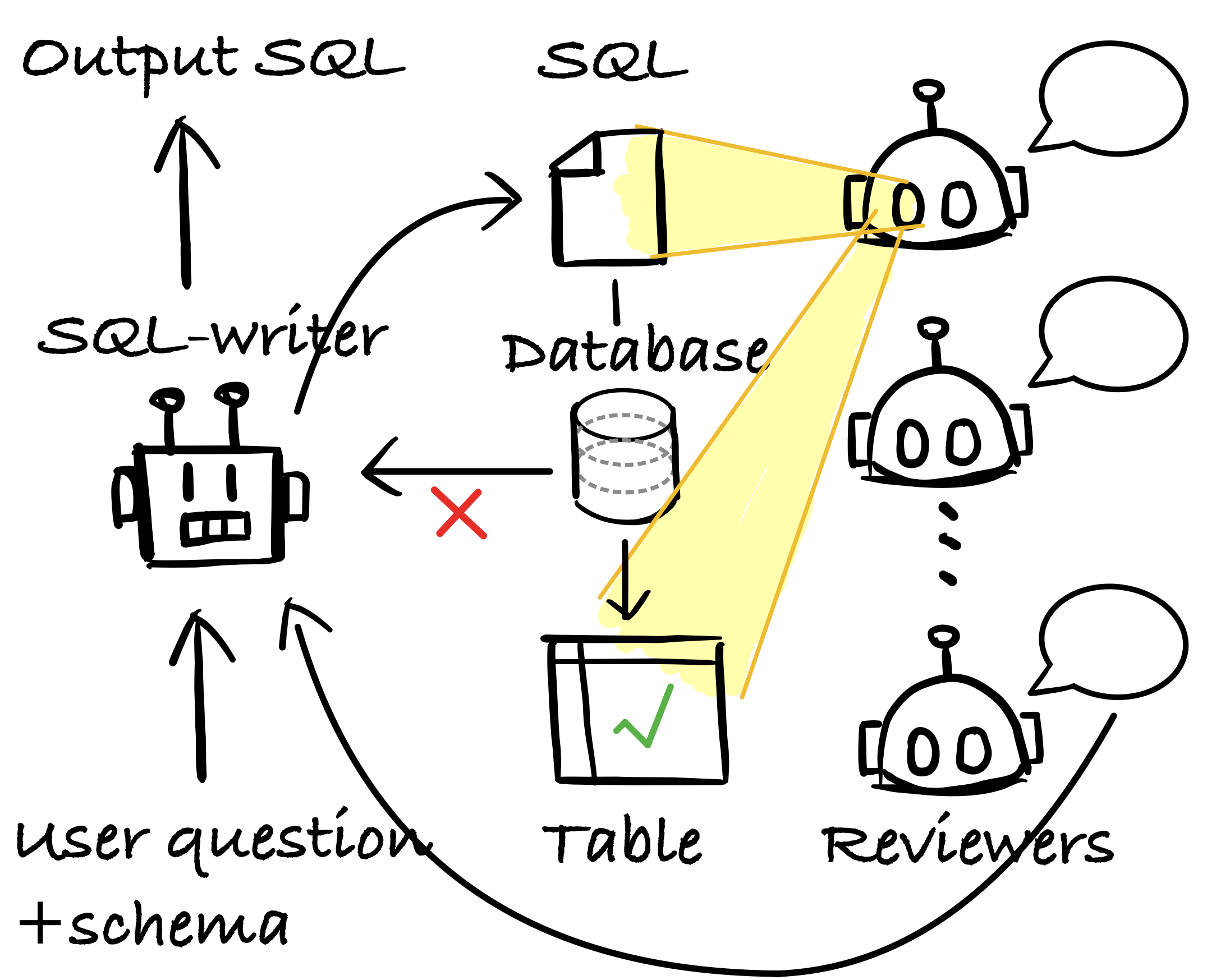}\\
    \caption{\sysname~ Architecture. $n$ reviewer agents, each with distinct characteristics, are created to review the generated SQL and its execution result. The process continues until the master node (SQL-writer) and the other nodes reach a consensus, at which point the system outputs the final SQL.}
    \label{fig:3r}
\end{figure}
We propose \sysname, a consensus-based multi-agent system for Text-to-SQL tasks. 
The proposed system draws inspiration from the peer-review mechanism, featuring one agent as the SQL-writer and several reviewers automatically generated by the LLM.
Once the generated SQL query is tested to be executable, the system will step into a review process, where we use the execution results to guide the SQL-writer and reviewers to refine the SQL. 
Through rounds of ``review'', ``negotiation or rebuttal'', and ``revision'', SQL-writer and reviewers will finally achieve consensus and deliver a solution with collective agreement (see \Cref{fig:3r}).

We test \sysname~on the popular Spider and Bird benchmarks.
\sysname~outperforms the existing single LLM as well as the multi-agent Tex-to-SQL systems by $1.3\%$ to $8.1\%$ on Spider and Bird. 
Surprisingly, we find that for Llama-3-8B, \sysname~outperforms chain-of-thought prompting by over 20\%,
even outperforming GPT-3.5 on the Spider-Dev set.
Our contributions can be summarized as follows:
\begin{enumerate}[leftmargin=\parindent,align=left,labelwidth=\parindent,labelsep=0pt]
    \item To the best of our knowledge, \sysname~is the first Text-to-SQL system to use the execution result for SQL refinements, and the first Text-to-SQL system to equip agents with memory sequences to enhance SQL generation.
    \item \sysname~offers a consensus-based multi-agent system for Text-to-SQL tasks. 
    Using very succinct prompts, it achieves strong performance compared to other systems.
    In addition, it effectively helps open-source LLMs such as Llama-3-8B on SQL generation.
    \item We provide a detailed error analysis of \sysname~on the existing Text-to-SQL benchmarks, shedding light on future research on the Text-to-SQL task.
\end{enumerate}



\section{Architecture}
\label{sec:system}

\paragraph{SQL-Writer (SW).} We task SW agents to: (1) compose the original SQL query based on the user question and database schema; (2) ensure that the SQL query is executable, and correct it when errors occur; (3) respond to reviewer agents' feedback and revise the SQL query accordingly. 
Specifically, we prompt SW agent through Prompt 1 in Appendix \ref{app:prompts}. 
For task (1), we feed the Prompt 1 to SW agent directly.
Given a user question $x$ and the database schema $\mathcal{S}$, task (1) can be formalized as:
\begin{equation*}
\setlength\abovedisplayskip{3pt}
\setlength\belowdisplayskip{3pt}
    y = {\rm \texttt{LLM}} (x, \mathcal{S}),
\end{equation*}
where $y$ is the generated SQL query.
For (2) and (3), we maintain a dialogue history $\mathcal{H}$ initially set to $\mathcal{H}=[(x, \mathcal{S}), y]$. Specifically, if an error $e$ occurs, we append $e$ to the history $\mathcal{H} \leftarrow \mathcal{H}+e$ and get $y'$ through:
\begin{equation*}
\setlength\abovedisplayskip{3pt}
\setlength\belowdisplayskip{3pt}
    y' = {\rm \texttt{LLM}} (\mathcal{H}).
\end{equation*}
We then concatenate $y'$ with the history $\mathcal{H}\leftarrow \mathcal{H}+y'$. 
In addition, considering the length limitation of LLMs' context window, we truncate the history $\mathcal{H}$ when the prompt is longer than the context limit.

\paragraph{Reviewers (REs).} 
We generate the reviewer agent's professions using an LLM (see Prompt 3 in Appendix \ref{app:prompts}) based on the database schema and the content of the SQL query, for instance, ``Senior Database Engineer specialized in writing various clauses'' and ``Data Analyst in the automotive industry'', etc. 
We incorporate these professions in the system prompt for the reviewer agent to make them focus on different aspects of the SQL query.
These reviewer agents are prompted to provide their professional comments based on the database schema, the user's question, the predicted SQL, and its execution result in the table format.

\paragraph{Overall Architecture.}
After several rounds of ``negotiation'' between the SQL-writer and reviewer agents, we decide whether there is a consensus by checking if the SQL-writer agent generates the same SQL query as in the previous round.
When there is a consensus, we terminate the negotiation loop and output the final SQL query. 
Algorithm \ref{algo:3r-loop} depicts the overall process of our system. 
\begin{algorithm}[t]
\small
\setlength\belowcaptionskip{-15pt}
\pmb{Given} $x$ (user question), $\mathcal{S}$ (schema)\\
$y = {\rm \texttt{LLM}} (x, \mathcal{S})$ \\
$\mathcal{H}=[(x, \mathcal{S}), y]$ \\
$i=0$\\
$j=0$\\
\While{$i<=\rm{MaxReviewTurns}$}{
    \While{$j<=\rm{MaxDebugTurns}$}{
        \textbf{Try:} \\
        \quad\quad $\mathcal{T}={\rm \texttt{Database}}(y)$ \\
        \quad\quad \textbf{break}\\
        \textbf{Except} \text{Exception as} $e$: \\
        $j \leftarrow j+1$ \\
        \quad\quad $\mathcal{H} \leftarrow \mathcal{H}+e$;\quad $y' = {\rm \texttt{LLM}} (\mathcal{H})$ \\
        \quad\quad $\mathcal{H} \leftarrow \mathcal{H}+y'$ \\
    }
    $r = {\rm \texttt{LLM}} (x, \mathcal{S}, y, \mathcal{T})$ \\
    $\mathcal{H} \leftarrow \mathcal{H}+r$;\quad $y'' = {\rm \texttt{LLM}} (\mathcal{H})$ \\
    $\mathcal{H} \leftarrow \mathcal{H}+y''$ \\
    \uIf{$y==y''$}{
        \textbf{break}}
    \Else{
        $y \leftarrow y''$}
    $i \leftarrow i+1$
}
\caption{\sysname-Loop}
\label{algo:3r-loop}
\end{algorithm}

\Cref{app:prompts} provides the detailed prompts we use in \sysname. 
In addition, we incorporate:
\begin{enumerate}[leftmargin=\parindent,align=left,labelwidth=\parindent,labelsep=0pt]
    \item Program of Thoughts (PoT) \cite{chen2022program} to prompt the SQL-writer agent to generate Python code before SQL query (see Prompt 2 in Appendix \ref{app:prompts}).
    Therefore, the agents may leverage Python in their reasoning process for better SQL query generation.
    \item $k$-shots example selection based on similarity of the user question embeddings.
    Specifically, when our system infers the SQL query in the test set, we select the $k$ most similar use questions and their corresponding SQL queries from the training set ($k$-shots) and use them for in-context learning.
\end{enumerate}

\section{Experiments and Results}
We conduct experiments on two cross-domain Text-to-SQL benchmarks, Spider and Bird detailed in \Cref{tab:benchmarks} in \Cref{app-sec:dataset-description}. 
We employ test-suite execution evaluation\footnote{github.com/taoyds/test-suite-sql-eval} \cite{zhong2020semantic}, the standard evaluation protocol for Spider, and the official SQL execution accuracy evaluation for Bird\footnote{bird-bench.github.io/}.
\Cref{tab:system-result} compares \sysname's performance with existing baseline methods when we employ different foundation LLMs.
Our best performed system achieves 88.1\%, 89.9\%, and 61.8\% on the Spider-Dev, Spider-Test, and Bird-Dev respectively, surpassing the existing multi-agent Text-to-SQL systems. 
In addition, we test our system with open-source Llama-3 models on Spider and report the results in \Cref{tab:llama-result}.
To our surprise, with the help of \sysname, zero-shot Llama-3-8B outperforms GPT-3.5 performance reported by \citet{li2023can} on Spider-Dev set.
This demonstrates the effectiveness of our proposed \sysname~system.


\setlength\tabcolsep{5pt}
\begin{table}[t]\small
    \centering
    \begin{tabular}{l|p{3.1cm}lll}
    \toprule
     Model  & Method          & SD & ST & BD  \\ \midrule
     \multicolumn{1}{c|}{\multirow{4}{*}{\rotatebox{300}{GPT-3.5}}} & - \cite{li2023can} & 72.1 & -        & 37.22          \\ 
            & C3 \cite{dong2023c3}            & 81.8           & 82.3          & -              \\
            & MAC \cite{wang2023mac}             & 80.6           & 75.5          & 50.56          \\ 
            & \sysname~ 5-shot & 81.4           & 81.1          & 52.15          \\ \midrule
     \multicolumn{1}{c|}{\multirow{5}{*}{\rotatebox{300}{GPT-4}}}  
            & DAIL \cite{gao2023text} & 83.6 & 86.6         & -              \\ 
            & PET \cite{li2024pet}            & 82.2           & 87.6          & -              \\ 
            & DIN \cite{pourreza2023din}            & 82.8           & 85.3          & 50.72          \\ 
            & MAC \cite{wang2023mac}            & 86.8           & 82.8          & 59.39          \\ 
            & \sysname~5-shot & \textbf{88.1}  & \textbf{89.9} & \textbf{61.80}
     \\ \bottomrule
    \end{tabular}
    \caption{Execution accuracy across various models and methods. We use the GPT-3.5-Turbo in our experiment. ``SD'', ``ST'', ``BD'' represent Spider-Dev, Spider-Test, Bird-Dev, respectively. For detailed description of baseline methods mentioned above, see Appendix \ref{app:baseline}.}
    \label{tab:system-result}
\end{table}

\begin{table}[t]\small
    \centering
    \begin{tabular}{llll}
    \toprule
     Model  & Method          & SD & ST  \\ \midrule
     GPT-3.5 & - \cite{li2023can} & 72.1 & -            \\ 
     \multicolumn{1}{l}{\multirow{2}{*}{Llama-3-8B}}  & CoT              & 52.1 & 53.5             \\ 
                 & \sysname~ 0-shot & 72.8 & 72.6              \\ 
     Llama-3-70B & \sysname~ 0-shot & 79.7 & 80.3              \\ \bottomrule
    \end{tabular}
    \caption{Execution accuracy comparison when we equip Llama-3 models with \sysname on Spider-Dev (``SD'') and Spider-Test (``ST'').}
    \label{tab:llama-result}
\end{table}

\subsection{Ablation Studies}
\definecolor{ColorSpider}{HTML}{FFC0CB}
\definecolor{ColorBird}{HTML}{87CEEB}
\newcommand{\applyGradient}[2]{
 \ifdefstring{#2}{--}{%
    #2
    }{
    \def\colorName{}
    \ifstrequal{#1}{Spider}{%
        \def\colorName{ColorSpider}
        \pgfmathsetmacro{\Lightness}{(#2-77.2)/(88.1-78.2)*200}
    }{%
        \def\colorName{ColorBird}
        \pgfmathsetmacro{\Lightness}{max((#2-54.2)/(61.8-37.2)*400, 0}
    }
    \edef\temp{\noexpand\cellcolor{\colorName!\Lightness!white}}\temp
    #2
    }
}
\setlength\tabcolsep{3pt}
\renewcommand{\arraystretch}{1.5}
\begin{table}[t]\small
    \centering
    \begin{tabular}{rcccc}
    \toprule
                  & \multicolumn{2}{c}{GPT-3.5-Turbo} & \multicolumn{2}{c}{GPT-4} \\
                  \cmidrule(lr){2-3}   \cmidrule(lr){4-5}
                  & Spider & Bird      & Spider   & Bird  \\ \midrule
    CoT           & 78.2   & 37.22     & 79.7     & 53.30 \\
    PoT           & 78.5   & 36.96     & 80.0     & 54.61 \\
    1R-Lp + CoT   & 78.3   & 44.13     & 82.3     & 57.89 \\
    1R-Lp + PoT   & 79.3   & 46.35     & 85.4     & 58.34 \\
    \textbf{\sysname: 3R-Lp + PoT}   & \textbf{81.4}   & \textbf{52.15}     & \textbf{88.1}     & \textbf{61.80} \\ \midrule
    \end{tabular}
    \caption{Ablation Studies on Spider-Dev and Bird-Dev (Execution Accuracy).
    The 1-Reviewer Loop (1R-Lp) represents that only one reviewer agent participates in the discussion, while the 3-Reviewers Loop (3R-Lp) represents three in the dicussion, which is also the default configuration of \sysname.
    We conduct all the experiments here under the 5-shot setting.}
    \label{tab:abl}
\end{table}
We conduct an ablation study on the impact of CoT, PoT with one or three reviewer agents in the discussion and report the results in \Cref{tab:abl}.
The results in Table \ref{tab:abl} show that the $n$-Reviewer(s) Loop ($n$R-Lp) plays a major role in performance improvement, with the 3R-Lp configuration significantly outperforming the 1R-Lp setup. The proposed \sysname~system achieves a 10.54\% improvement over the baseline GPT-4 + CoT. 
We provide the statistical significant test for these results in \Cref{app:significance}.
\Cref{app:k-shots} provides a sensitivity analysis of the impacts of the $k$ value in $k$-shots.

We conducted case studies on 244 instances from the Spider-Dev dataset where the CoT fail but \sysname succeed when combined with Llama-3-8B. The findings are as follows: 
\begin{enumerate}
    \item \textbf{Corrected non-executable SQL queries, 51\%.} The LLM equipped with memory module (see Section \ref{sec:system}) excels at correcting non-executable SQL queries.
    \item \textbf{Refinement based on reviewers' comments, 44\%.} The $n$R-Lp functions as an enhanced Self-Consistency (SC) \cite{wang2022self}. On the one hand, it avoids the hallucinations caused by high temperatures \cite{renze2024effect}. On the other hand, the $n$R-Lp considers feedback from all agents, unlike the voting process in SC, which consistently disregards minority opinions.
    \item \textbf{Refinement based on the output table, 27.5\%.} LLMs may not experts in SQL writing, but they are full-skilled data readers. The information that reviewers observe from execution results greatly assists the SQL-writer in refining the SQL.
\end{enumerate}

\begin{table*}[ht]
    \includegraphics[width=\linewidth]{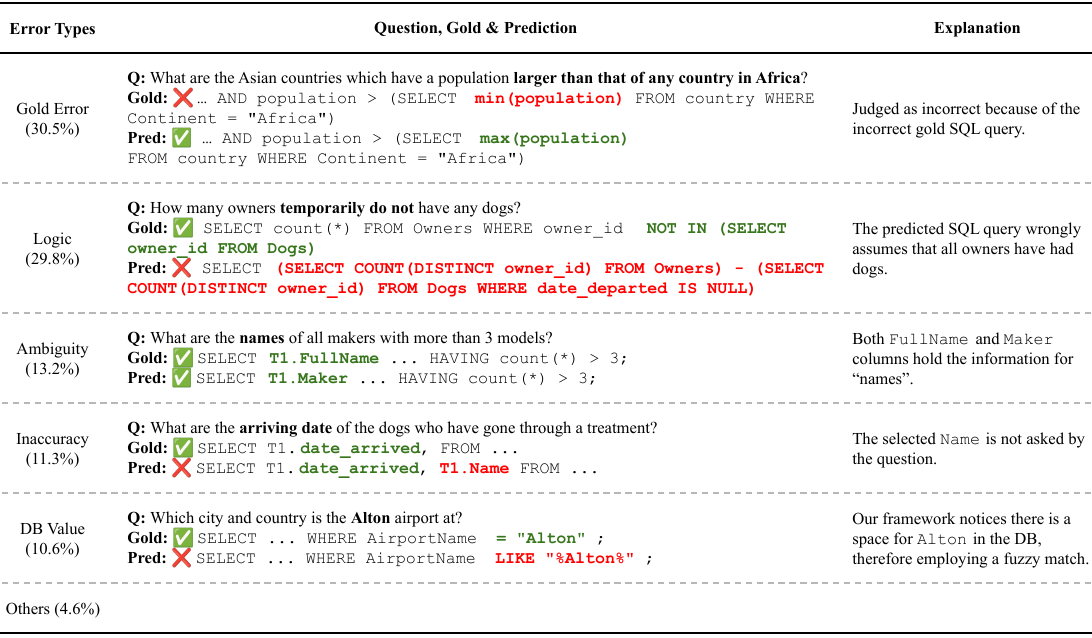}
    \caption{Error Analysis of \sysname~ on Spider-Dev.
    We make the part in the question red when it is either annotated incorrectly in the gold SQL query (Gold) or predicted incorrectly in the predicted SQL query (Pred).}
    \label{tab:error-analysis}
\end{table*}
\subsection{Error Analysis}
\label{sec: error-analysis}
In total, GPT-4+\sysname~ fails to generate the gold SQL queries for 123 instances in Spider-Dev.
\Cref{tab:error-analysis} shows the error case distribution for our system on Spider-Dev (more in \Cref{app:spider-error-cases,app:bird-error-cases}). Note that though we have spotted issues with the gold SQL queries, we still adopt the original set to calculate the performance of our system to ensure a fair comparison. 

\paragraph{Gold Error.}
We notice that though the annotation quality of Spider is good, there are still cases where the gold SQL queries are not correct.
Specifically, among the 151 examples, 30.5\% are due to incorrect gold SQL queries (4.5\% of all the examples in Spider-Dev). 
To facilitate future research, we catalog the instances with incorrect gold SQL, correct the errors, and share the details\footnote{visible-after-review.com}.

\paragraph{Ambiguity.}
We observe that there are a few questions involving ambiguities, a phenomenon spotted on a wide range of NLP tasks \cite{plank-2022-problem, deng-etal-2023-annotate}. 
In \Cref{tab:error-analysis}.3, both \texttt{FullName} and \texttt{Maker} columns hold the information for the ``name of makers'', except that \texttt{FullName} holds the full names while \texttt{Maker} holds the name abbreviations.
Therefore, both the gold and predicted SQL queries should be considered correct if there is no further clarifications.
Such ambiguous requests may be common in real-world applications as the lay users may not be familiar with the database schema.
This requires future research on interactive Text-to-SQL systems that can understand and deal with such ambiguities in user questions.

\paragraph{Dirty Database Value.}
We observe that due to the Database (DB) setup for Spider, certain DB values may deviate from what is asked in the question.
For instance, in \Cref{tab:error-analysis}.5, \sysname~notices a space for \texttt{Alton} in DB, therefore employing a fuzzy match.
But this deviates the SQL query's execution results from the gold SQL query's results.

Explanations of ``Logic'' and ``Inaccuracy'' errors can be found in Appendix \ref{app:add-error-types}. Our findings indicate that the existing evaluation protocols for Text-to-SQL generation may not authentically capture the capabilities of these sophisticated systems. 
Therefore, we advocate for a reassessment and enhancement of Text-to-SQL evaluation methods. 
We provide further error analysis of \sysname~on Bird in \Cref{app:bird-error-cases}.



\section{Conclusion}
\label{sec: conclusion}
\sysname~significantly enhance the performances of LLMs on the Text-to-SQL task.
We conduct a comprehensive error analysis and identify persistent issues with the current Text-to-SQL evaluation. 
This underscores the necessity for our community to develop a refined evaluation protocol that more effectively captures nuances in SQL generation and accurately reflects model performance.

\section*{Limitations}
Due to the scope of the study, we only test a limited number of LLMs.
The performance gap between 1R-Lp and 3R-Lp demonstrates that the number of reviewers is a worthwhile topic of research. However, this work does not delve into this much.


\section*{Ethical Statements}
In this paper, we propose strategies to improve the SQL generation capabilities of LLMs.
To the best of our knowledge, we do not expect our system would have negative impacts on the society.

\bibliography{custom}

\appendix
\section{Appendix}
\label{app}

\subsection{Dataset Descriptions}
\label{app-sec:dataset-description}

\setlength\tabcolsep{4pt}
\begin{table}[h]\small
    \centering
    \begin{tabular}{lllr}
        \toprule
             & Spider-Dev & Spider-Test & Bird-Dev  \\
             & \multicolumn{2}{c}{\cite{yu2018spider}} & \cite{li2023can} \\
        \midrule
        \#QA     & 1,034  &2147& 1,534\\
        \#Domain & 138    & -& 37 \\
        \#DB     & 200    &206& 95 \\
        DB Size  & 879.5 MB  &906.5 MB & 1.76 GB \\
        \bottomrule
    \end{tabular}
    \caption{Statistics of two Text-to-SQL benchmarks we use in our experiments. ``\#QA'', ``\#Domain'' and ``\#DB'' refer to the number of samples, domains and databases, respectively.}
    \label{tab:benchmarks}
\end{table}

\subsection{Baseline}
\label{app:baseline}
Experiments in this work was based on LLMs including GPT-3.5-Turbo, GPT-4 \cite{openai2023gpt} and Llama-3 \cite{llama3modelcard}. As for the compared methods, the raw performance for GPT-3.5 (``-'') was evaluated by \citet{li2023can}; C3 employs schema linking filtering \cite{dong2023c3}; DAIL selects few-shot demonstrations based on their skeleton similarities \cite{gao2023text}, and ``SC'' represents Self-Consistency \cite{wang2022self}; PET uses cross-consistency \cite{li2024pet}; DIN decomposes the text-to-SQL task into smaller subtasks \cite{pourreza2023din}; MAC, as previously mentioned, is the first to apply a Multi-Agent system to Text-to-SQL tasks \cite{wang2023mac}.

\subsection{Effects of $k$ in $k$-shot.}
\label{app:k-shots}
\begin{figure}[ht]
\centering
    \includegraphics[width=.9\columnwidth]{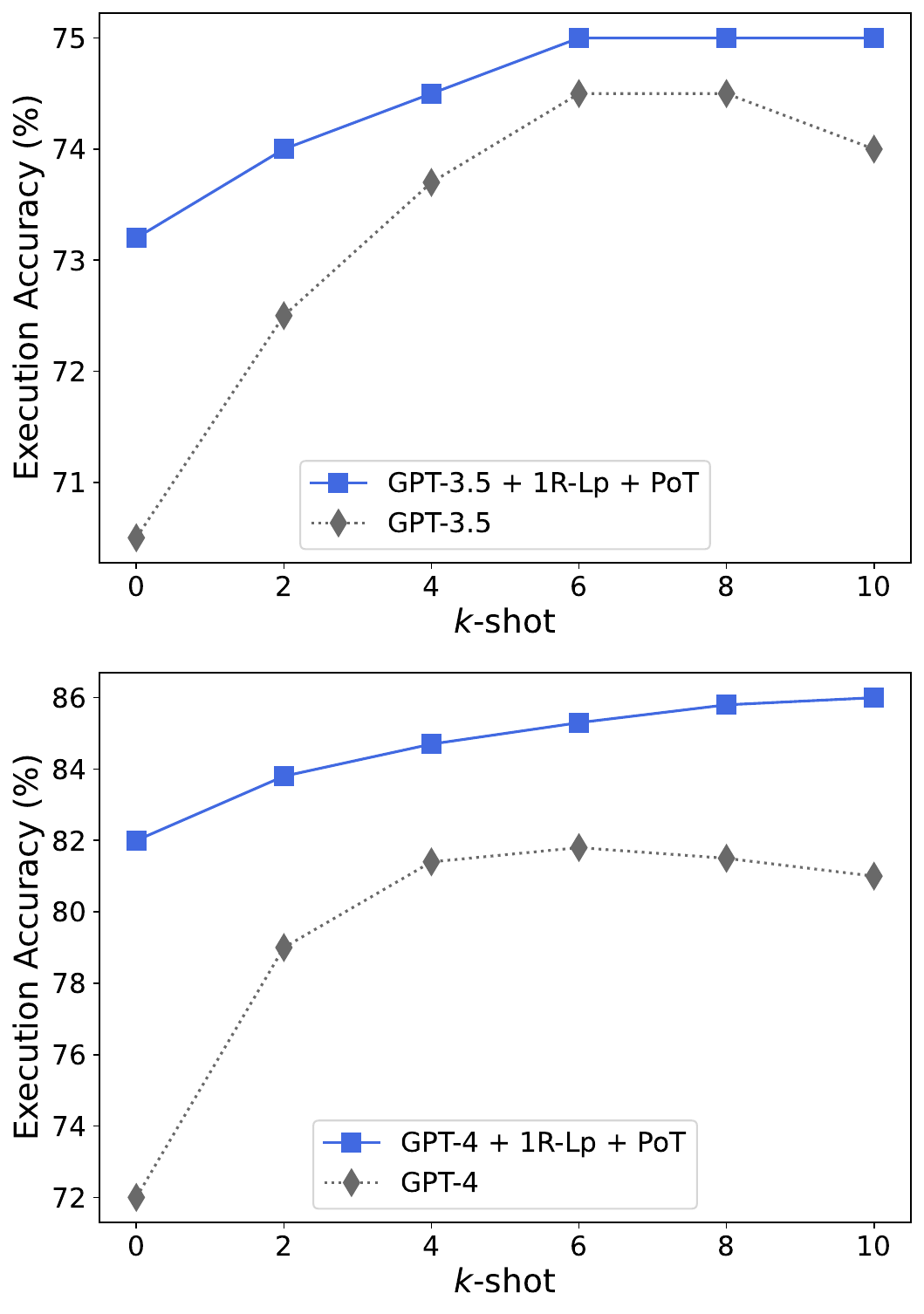}
    \caption{$k$-shot Sensitivity Analysis. 
}
\label{fig:few-shots}
\end{figure}

We test various $k$ values on 200 random samples from Spider-Dev. 
As shown in Figure \ref{fig:few-shots}, compared to CoT, the performance of the \sysname~system remains relatively stable regardless of the number of examples, which corroborates our previous findings from the 0-shot experiments with Llama-3.

\subsection{Significance Test}
\label{app:significance}
We divided the generated SQL by several strategies in Table \ref{tab:abl} into 10 equal parts and calculated the execution accuracy for each. To test whether our strategy can indeed improve execution accuracy, we conduct a significance test between the ``CoT'' and ``3R-Lp+PoT'' strategies. The null hypothesis of the test is that the median execution accuracy obtained by the two strategies is the same. The Mann-Whitney U Test \cite{mann1947test} is a non-parametric statistical method used to compare whether there is a significant difference in the medians of two independent samples. Compared to the Analysis of Variance (ANOVA), it does not require the data to be normally distributed, making it suitable for small samples or data with unknown distribution.

The $p$-value of the test is 0.0024, which is below the commonly accepted significance level of 0.05. Therefore, we have reason to reject the null hypothesis, indicating that the ``3R-Lp+PoT'' strategy leads to a significant performance improvement.

\subsection{Additional Error Types}
\label{app:add-error-types}
\paragraph{Logic.}
In \Cref{tab:error-analysis}.2, we present an example of the logic error made by \sysname.
We notice that LLMs may solve the problems using a more complicated logic, which is prone to mistakes.
For instance, in \Cref{tab:error-analysis}.2, instead of directly counting the owners who do not own dogs, the LLMs try to substract the number of dog owners from the total number of owners.
This ignores the possibility that some owners may have never had any dogs before.
This addresses an issue with the multi-agent system that if the system comes up with a complicated initial SQL query, the following discussion process may try to polish the complicated SQL query instead of switching to an easier solution.
In cases like \Cref{tab:error-analysis}.2, there is no way to reach a perfect SQL query with the substraction logic.

\paragraph{Inaccuracy.}
We observe that the LLMs may incorporate more information than what is asked by the end user.
For instance, in \Cref{tab:error-analysis}.4, the user does not ask for the name of the dogs but the LLMs present such information along with the asked arriving date.
We hypothesize that since such extra information can potentially be helpful to the end user, LLMs may be biased towards including it.



\onecolumn
\newpage

\subsection{Spider Error Cases}
\label{app:spider-error-cases}
\definecolor{gold}{HTML}{98FB98}
\definecolor{pred}{HTML}{FFC0CB}
\newcolumntype{L}[1]{>{\raggedright\arraybackslash}p{#1}}
\begin{table}[h!]\small
    \centering
    \begin{tabular}{lL{8cm}L{4cm}}
    \toprule
        Error Type & Question, Gold \& Prediction & Reason  \\ \toprule 
        
        DB Value & 
        \pmb{Q:} Find the last name of the students who currently live in the state of North Carolina but have not registered in any degree program. \newline
        \pmb{Gold:} \texttt{SELECT ... WHERE T2.state\_province\_county \colorbox{gold}{= 'NorthCarolina'} EXCEPT ...} \newline 
        \pmb{Pred:} \texttt{SELECT ... WHERE T2.state\_province\_county \colorbox{pred}{= 'North Carolina'} EXCEPT ...} &
        The filtering condition in the question does not match the database value, string ``NorthCalifornia'' in database do not have a space in between. \\ \hline
        
        
        Gold Error & 
        \pmb{Q:} What are the first names of all players, and their average rankings? \newline
        \pmb{Gold:} \texttt{SELECT avg(ranking), T1.first\_name FROM players AS T1 JOIN rankings AS T2 ON T1.player\_id = T2.player\_id \colorbox{pred}{GROUP BY T1.first\_name}} \newline 
        \pmb{Pred:} \texttt{SELECT avg(ranking), T1.first\_name FROM players AS T1 JOIN rankings AS T2 ON T1.player\_id = T2.player\_id \colorbox{gold}{GROUP BY T1.player\_id}} &
        The individuals in the table can be uniquely determined by column player\_id not first\_name, when \texttt{GROUP BY}. \\ \hline
        
        Gold Error &
        \pmb{Q:} Find the id and cell phone of the professionals who operate two or more types of treatments.\newline
        \pmb{Gold:} \texttt{SELECT T1.professional\_id, T1.cell\_number FROM Professionals AS T1 JOIN Treatments AS T2 ON T1.professional\_id = T2.professional\_id GROUP BY T1.professional\_id \colorbox{gold}{HAVING count(*) >= 2}} \newline
        \pmb{Pred:} \texttt{SELECT T1.professional\_id, T1.cell\_number FROM Professionals AS T1 JOIN Treatments AS T2 ON T1.professional\_id = T2.professional\_id GROUP BY T1.professional\_id HAVING COUNT(\colorbox{pred}{DISTINCT} T2.treatment\_type\_code) >= 2} &
        The gold only finds professionals who have two or more records in the treatment table does not ensure that the records are for different types of treatments \\ \hline
        
        Ambiguity &  
        \pmb{Q:} What are the names and ids of all makers with more than 3 models? \newline
        \pmb{Gold:} \texttt{SELECT \colorbox{gold}{T1.FullName}, T1.Id FROM CAR\_MAKERS AS T1 JOIN MODEL\_LIST AS T2 ON T1.Id  =  T2.Maker GROUP BY T1.Id HAVING count(*) > 3;} \newline
        \pmb{Pred:} \texttt{SELECT \colorbox{pred}{T1.Maker}, T1.Id FROM CAR\_MAKERS AS T1 JOIN MODEL\_LIST AS T2 ON T1.Id  =  T2.Maker GROUP BY T1.Id HAVING count(*) > 3;} &
        Both column ``Maker'' and column ``FullName'' can answer the question about the ``names of makers'' in the query.\\ \hline
        
        Imprecise & 
        \pmb{Q:} What are the arriving date and the departing date of the dogs who have gone through a treatment?\newline
        \pmb{Gold:} \texttt{SELECT DISTINCT T1.date\_arrived, T1.date\_departed FROM Dogs AS T1 JOIN Treatments AS T2 ON T1.dog\_id = T2.dog\_id} \newline 
        \pmb{Pred:} \texttt{SELECT DISTINCT T1.date\_arrived, T1.date\_departed, \colorbox{pred}{T1.Name} FROM Dogs AS T1 JOIN Treatments AS T2 ON T1.dog\_id = T2.dog\_id} &
        The question do not require listing the specific names of the students, but only ask to list the students' arrival and departure dates. This falls under information redundancy.\\ 
        
    \bottomrule
    \end{tabular}
    \caption{}
\end{table}

\newpage
\subsection{BIRD Error Cases}
\label{app:bird-error-cases}

\definecolor{gold}{HTML}{98FB98}
\definecolor{pred}{HTML}{FFC0CB}
\newcolumntype{L}[1]{>{\raggedright\arraybackslash}p{#1}}
\begin{table}[h!]\small
    \centering
    \begin{tabular}{lL{8cm}L{4cm}}
    \toprule
        Error Type & Question, Gold \& Prediction & Reason  \\ \toprule 
        
        DB Value & 
        \pmb{Q:} How many cards with unknown power that can't be found in foil is in duel deck A? \newline
        \pmb{Gold:} \texttt{SELECT SUM(CASE WHEN power \colorbox{gold}{LIKE '\%*\%'} OR power IS NULL THEN 1 ELSE 0 END) FROM cards WHERE hasFoil = 0 AND duelDeck = 'a'} \newline 
        \pmb{Pred:} \texttt{SELECT COUNT(*) FROM cards WHERE (power IS NULL OR \colorbox{pred}{power = '*'}) AND hasFoil = 0 AND duelDeck = 'a'} &
        Values in database cannot exact match with the question and evidence. \\ \hline

        Gold Error &
        \pmb{Q:} How many artists have designed a card with a black border color and is available in both "arena" and "mtgo" printing type? \newline
        \pmb{Gold:} \texttt{SELECT COUNT(CASE WHEN availability LIKE '\%arena,mtgo\%' THEN 1 ELSE NULL END) FROM cards} \newline
        \pmb{Pred:} \texttt{SELECT COUNT(DISTINCT artist) FROM cards \colorbox{gold}{WHERE borderColor = 'black'} AND availability LIKE '\%arena,mtgo\%'} &
        The gold missed one filtering condition. \\ \hline
        
        Ambiguity &  
        \pmb{Q:} Among black card borders, which card has full artwork? \newline
        \pmb{Gold:} \texttt{SELECT \colorbox{gold}{id} FROM cards WHERE borderColor = 'black' AND isFullArt = 1} \newline
        \pmb{Pred:} \texttt{SELECT \colorbox{pred}{name} FROM cards WHERE borderColor = 'black' AND isFullArt = 1;} &
        Both column ``name'' and column ``id'' can answer the question about the ``the card has full artwork'' in the query.\\ \hline

        Logic & 
        \pmb{Q:} Which cities have the top 5 lowest enrollment number for students in grades 1 through 12? \newline
        \pmb{Gold:} \texttt{SELECT T2.City FROM frpm AS T1 INNER JOIN schools AS T2 ON T1.CDSCode = T2.CDSCode GROUP BY T2.City ORDER BY SUM(T1.`Enrollment (K-12)`) ASC LIMIT 5} \newline 
        \pmb{Pred:} \texttt{SELECT T2.City, T1."Enrollment (K-12)"  FROM frpm AS T1  JOIN schools AS T2  ON T1.CDSCode = T2.CDSCode  ORDER BY T1."Enrollment (K-12)" ASC  LIMIT 5;} &
        The Pred query is incorrect because it lacks grouping and aggregation. It directly selects cities and their corresponding "Enrollment (K-12)" without summing up the enrollment numbers across all schools within each city. As a result, it returns the cities of the five schools with the lowest enrollment numbers, which does not accurately answer the question seeking the top 5 cities with the lowest total enrollment for students in grades 1 through 12.\\ 

    \bottomrule
    \end{tabular}
    \caption{}
\end{table}

\newpage
\subsection{Interactive Correction Cases}
\label{app:ic-cases}

\definecolor{gold}{HTML}{98FB98}
\definecolor{pred}{HTML}{FFC0CB}
\newcolumntype{L}[1]{>{\raggedright\arraybackslash}p{#1}}
\begin{table}[h!]\small
    \centering
    \begin{tabular}{L{5cm}|L{5cm}|L{5cm}}
    \toprule
        Before & After & Reason  \\ \toprule 
        \multicolumn{3}{L{15cm}}{\pmb{Q:} What is the phone number of the school that has the highest number of test takers with an SAT score of over 1500?} \\ \hline
        \pmb{Before:} \newline
        \texttt{
        SELECT schools.Phone \newline
        FROM satscores \newline
        JOIN schools ON satscores.cds = schools.CDSCode \newline
        WHERE satscores.NumGE1500 > 1500 \newline
        GROUP BY schools.School \newline
        ORDER BY COUNT(*) DESC \newline
        LIMIT 1;} &
        \pmb{After:} \newline
        \texttt{
        SELECT schools.Phone \newline
        FROM satscores \newline
        JOIN schools ON satscores.cds = schools.CDSCode \newline
        ORDER BY satscores.NumGE1500 DESC \newline
        LIMIT 1;} &
     The WHERE clause is checking if the number of test takers with an SAT score over 1500 is greater than 1500, which is not what the question is asking. The question is asking for the school with the highest number of test takers who scored over 1500.   Additionally, the GROUP BY clause is grouping by the school name, which may not be unique. It would be better to group by the school code (CDSCode) which is unique for each school. \\ \bottomrule
     
        \multicolumn{3}{L{15cm}}{\pmb{Q:} List the top five schools, by descending order, from the highest to the lowest, the most number of Enrollment (Ages 5-17). Please give their NCES school identification number.} \\ \hline
        \pmb{Before:} \newline
        \texttt{
        SELECT T1."School Name", T2.NCESSchool \newline
        FROM frpm AS T1 \newline
        JOIN schools AS T2 ON T1.CDSCode = T2.CDSCode \newline
        ORDER BY T1."Enrollment (Ages 5-17)" DESC \newline
        LIMIT 5;} &
        \pmb{After:} \newline 
        \texttt{
        SELECT T2.NCESSchool \newline
        FROM frpm AS T1 \newline
        JOIN schools AS T2 ON T1.CDSCode = T2.CDSCode \newline
        ORDER BY T1."Enrollment (Ages 5-17)" DESC \newline
        LIMIT 5;} &
    The question asks for the top five schools with the most number of Enrollment (Ages 5-17), and their NCES school identification number. However, the SQL query provided returns the school name, which is not required by the question. \\ \bottomrule

        \multicolumn{3}{L{15cm}}{\pmb{Q:} find the package option of the tv channel that do not have any cartoon directed by Ben Jones.} \\ \hline
        \pmb{Before:} \newline
        \texttt{
        SELECT TV\_Channel.Package\_Option \newline
        FROM TV\_Channel \newline
        LEFT JOIN Cartoon ON TV\_Channel.id = Cartoon.Channel \newline
        WHERE Cartoon.Directed\_by IS NOT 'Ben Jones';}
        &
        \pmb{After:} \newline  
        \texttt{
        SELECT TV\_Channel.Package\_Option \newline
        FROM TV\_Channel \newline
        WHERE TV\_Channel.id NOT IN ( \newline
            SELECT Cartoon.Channel \newline
            FROM Cartoon \newline
            WHERE Cartoon.Directed\_by = 'Ben Jones');}&
    The current query is selecting all channels where the director is not Ben Jones, but it should exclude channels that have any cartoons directed by Ben Jones. This can be achieved by using a subquery to find all channels that have cartoons directed by Ben Jones, and then excluding these channels in the main query. Here is the corrected SQL: \\ \bottomrule

        \multicolumn{3}{L{15cm}}{\pmb{Q:} List the earnings of poker players in descending order.} \\ \hline
        \pmb{Before:} \newline
        \texttt{
        SELECT * FROM poker\_player ORDER BY Earnings DESC;}
        &
        \pmb{After:} \newline
        \texttt{
        SELECT Earnings FROM poker\_player ORDER BY Earnings DESC;} &
    The question only asks for the earnings of poker players. Therefore, we should only select the `Earnings` column to make the result more concise and relevant to the question.\\ \bottomrule

        \multicolumn{3}{l}{\pmb{Q:} How many countries speak both English and Dutch?} \\ \hline
        \pmb{Before:} \newline
        \texttt{
        SELECT COUNT(DISTINCT CountryCode) \newline
        FROM countrylanguage \newline
        WHERE Language = 'English' \newline
        AND CountryCode IN (\newline
            SELECT CountryCode \newline
            FROM countrylanguage \newline
            WHERE Language = 'Dutch')
        }
        &
        \pmb{After:} \newline
        \texttt{
        SELECT COUNT(*) \newline
        FROM (\newline
            SELECT CountryCode \newline
            FROM countrylanguage \newline
            WHERE Language = 'English' \newline
            INTERSECT \newline
            SELECT CountryCode \newline
            FROM countrylanguage \newline
            WHERE Language = 'Dutch')
        }  &
    The SQL query provided uses the `IN` clause, which selects countries that speak either English or Dutch. However, the question asks for countries speaking both languages. Using IN results in a union of sets (either English or Dutch), not an intersection (both English and Dutch). This could be done with two separate queries joined by an `INTERSECT` clause.\\ \bottomrule
    \end{tabular}
    \caption{}
    \label{}
\end{table}

\newpage
\subsection{Prompt Examples}
\label{app:prompts}

\lstset{
  language=Python,                
  upquote=true,
  basicstyle=\fontfamily{cmtt}\selectfont,     
  commentstyle=\normalfont,       
  backgroundcolor=\color{olive!10},
  frame=none,                     
  breaklines=true,                
  showstringspaces=false,         
  tabsize=2,                      
  breakindent=0pt,                
  captionpos=b,                   
  escapeinside={(*@}{@*)},
}

\def\specialcyan{\begingroup\color{cyan}\fontfamily{cmtt}}
\def\endspecialcyan{\endgroup}
\def\specialgreen{\begingroup\color{green}\fontfamily{cmtt}}
\def\endspecialgreen{\endgroup}

\begin{lstlisting}[title={Prompt 1: CoT-SQL-Writer}]
Describe how you understand the question based on the evidence, and help me write an SQL to answer the question.
### EVIDENCE: (*@\aftergroup\specialgreen@*){(*@\aftergroup\endspecialgreen@*)(*@\aftergroup\specialcyan@*)evidence(*@\aftergroup\endspecialcyan@*)(*@\aftergroup\specialgreen@*)}(*@\aftergroup\endspecialgreen@*)
### USER_QUESTION: (*@\aftergroup\specialgreen@*){(*@\aftergroup\endspecialgreen@*)(*@\aftergroup\specialcyan@*)question(*@\aftergroup\endspecialcyan@*)(*@\aftergroup\specialgreen@*)}(*@\aftergroup\endspecialgreen@*)

### RELATED SQL:
(*@\aftergroup\specialgreen@*){(*@\aftergroup\endspecialgreen@*)(*@\aftergroup\specialcyan@*)related_sql(*@\aftergroup\endspecialcyan@*)(*@\aftergroup\specialgreen@*)}(*@\aftergroup\endspecialgreen@*)

### DATABASE STRUCTURE:
(*@\aftergroup\specialgreen@*){(*@\aftergroup\endspecialgreen@*)(*@\aftergroup\specialcyan@*)schema(*@\aftergroup\endspecialcyan@*)(*@\aftergroup\specialgreen@*)}(*@\aftergroup\endspecialgreen@*)
\end{lstlisting}

\begin{lstlisting}[title={Prompt 2: PoT-SQL-Writer}]
Write an to answer the question.

Program of Thoughts (PoT) is a variant of Chain of Thought (CoT), pre-generating Python code to assist in the creation of SQL. Please apply PoT (and PoT only) before generating an SQL.
In your python code, `Table %s` is stored in `db_dict['%s']`, `db_dict` is of type dict[pandas.DataFrame].

### RELATED SQL:
(*@\aftergroup\specialgreen@*){(*@\aftergroup\endspecialgreen@*)(*@\aftergroup\specialcyan@*)related_sqls(*@\aftergroup\endspecialcyan@*)(*@\aftergroup\specialgreen@*)}(*@\aftergroup\endspecialgreen@*)

### DATABASE STRUCTURE:
(*@\aftergroup\specialgreen@*){(*@\aftergroup\endspecialgreen@*)(*@\aftergroup\specialcyan@*)schema(*@\aftergroup\endspecialcyan@*)(*@\aftergroup\specialgreen@*)}(*@\aftergroup\endspecialgreen@*)

### EXAMPLES:
QUESTION: What is %s in the earliest year and what year was it?
SQL:
earliest_year = db_dict[%s]['Year'].min()
year_filtered_data = step1_result[step1_result['Year'] == earliest_year]
result = year_filtered_data[[%s, 'Year']]
```sql
SELECT T1.%s, T2.Year FROM %s AS T1 JOIN %s AS T2 ON T1.Id = T2.Id WHERE T2.Year = (SELECT min(YEAR) FROM %s);
```

QUESTION: Show names for all %s except for %s having a %s in year 2023.
SQL:
%s_2023 = db_dict['%s'][db_dict['%s']['year'] == '2023']
result = db_dict[%s][~db_dict[%s][%s].isin(%ss_2023[%s])]
```sql
SELECT name FROM %s EXCEPT SELECT T2.name FROM %s AS T1 WHERE T1.year = 2023
```

QUESTION: Find the %s that %s is A and B?
SQL:
condition_a_data = db_dict[%s][db_dict['Cartoon'][%s] == 'A']
condition_b_data = db_dict[%s][db_dict['Cartoon'][%s] == 'B']
result = pd.merge(condition_a_data, condition_b_data, how='inner')
```sql
SELECT T1.%s FROM %s AS T1 WHERE %s = 'A' 
INTERSECT 
SELECT T1.%s FROM %s AS T1 WHERE %s = 'B' 
```

### EVIDENCE: (*@\aftergroup\specialgreen@*){(*@\aftergroup\endspecialgreen@*)(*@\aftergroup\specialcyan@*)evidence(*@\aftergroup\endspecialcyan@*)(*@\aftergroup\specialgreen@*)}(*@\aftergroup\endspecialgreen@*)
### USER_QUESTION: (*@\aftergroup\specialgreen@*){(*@\aftergroup\endspecialgreen@*)(*@\aftergroup\specialcyan@*)question(*@\aftergroup\endspecialcyan@*)(*@\aftergroup\specialgreen@*)}(*@\aftergroup\endspecialgreen@*)
### SQL:
\end{lstlisting}

\begin{lstlisting}[title={Prompt 3: Invitation}]
You are the manager of a Database project. You are going to invite (*@\aftergroup\specialgreen@*){(*@\aftergroup\endspecialgreen@*)(*@\aftergroup\specialcyan@*)n(*@\aftergroup\endspecialcyan@*)(*@\aftergroup\specialgreen@*)}(*@\aftergroup\endspecialgreen@*) experts to review an SQL query.
Who would you invite?

considering:
(1) the domain of this database; 
(2) the structure of this SQL.
Please write your invitation as a JSON format dictionary, Enclose the JSON within ```json...```.

### DATABASE STRUCTURE:
(*@\aftergroup\specialgreen@*){(*@\aftergroup\endspecialgreen@*)(*@\aftergroup\specialcyan@*)schema(*@\aftergroup\endspecialcyan@*)(*@\aftergroup\specialgreen@*)}(*@\aftergroup\endspecialgreen@*)

### QUESTION: (*@\aftergroup\specialgreen@*){(*@\aftergroup\endspecialgreen@*)(*@\aftergroup\specialcyan@*)question(*@\aftergroup\endspecialcyan@*)(*@\aftergroup\specialgreen@*)}(*@\aftergroup\endspecialgreen@*)
### SQL: 
(*@\aftergroup\specialgreen@*){(*@\aftergroup\endspecialgreen@*)(*@\aftergroup\specialcyan@*)pred_sql(*@\aftergroup\endspecialcyan@*)(*@\aftergroup\specialgreen@*)}(*@\aftergroup\endspecialgreen@*)

### EXAMPLES:
```json
{
  "Reviewer PVsg": "Data Analyst in automotive industry",
  "Reviewer 2KtR": "Senior Database Engineer specialized in writing various clauses",
  "Reviewer LmN3": "Senior Database Engineer specialized in writing filtering conditions"
}
```
### INVITATION:
\end{lstlisting}

\end{document}